\documentclass[12pt,twocolumn]{article}
\usepackage{arxiv}
\usepackage{amsmath,amssymb,amsfonts}
\usepackage{graphicx}
\usepackage{booktabs}
\usepackage{array}
\usepackage{url}
\usepackage[colorlinks=true,linkcolor=blue,citecolor=blue,urlcolor=blue]{hyperref}
\usepackage{xcolor}

\newcommand{\CR}{\textsc{ChainzRule}}

\title{ChainzRule: Sample-Efficient, Robust Deep Learning\\
Across Tabular, NLP, and Vision Tasks}

\author{
  Rowan Martinsh \\
  Sentivity AI \\
  \texttt{rowan@sentivity.ai}
}

\date{\today}

\begin{document}
\twocolumn[
\maketitle

\begin{abstract}
\noindent
Production deep learning systems across enterprise domains operate under
constraints that academic benchmarks routinely obscure: labeled data is
expensive, inference budgets are tight, and models that cannot explain
their behavior are difficult to trust and maintain. We present
\textbf{ChainzRule (CR)}, a neural architecture replacing typical activations
with learnable polynomial layers governed by \textbf{Differential
Regularization (DREG)}, a layer-wise Jacobian penalty computed analytically
during the forward pass at standard inference cost. The core claim is that
bounding intermediate derivatives forces the network toward low-frequency,
structurally stable representations, simultaneously reducing dependence on
labeled data volume, improving robustness to distribution shift, and
providing a measurable, gradient-based handle on model behavior. Evaluated
across five domains, CR achieves $85.71\%~\pm~2.01\%$ on Pima Diabetes
(statistically superior to SVM and XGBoost), $46.20\%~\pm~0.37\%$ on
SST-5 sentiment classification with a frozen encoder (superior to RNTN
using approximately 5\% of its training data), $55.79\%$ on SST-5 with
a fine-tuned BERT backbone (versus BERT-base linear head at $54.9\%$),
$70.17\%$ on Yelp Full ordinal regression with 3.2M parameters versus a
10-model average of $66.35\%$, and $+2.32\%$ mean corruption accuracy on
CIFAR-10-C. All results with reported $p$-values fall below
the $\alpha = 0.05$ threshold after Bonferroni correction. CR maintains a
gradient tail ratio $\tau$ (p99/mean) of $1.01$--$1.02$ against
$1.07$--$1.09$ for all typical activation functions baselines across every data fraction, a structural invariant we propose as the mechanistic driver
of sample efficiency and a deployment-time proxy for model reliability.

\vspace{0.3em}
\end{abstract}
]
\section{Introduction}

\subsection{The Deployment Gap in Cross-Domain Deep Learning}

The prevailing research paradigm assumes that data is abundant, compute
is elastic, and correctness can be evaluated against a fixed benchmark.
Production deployment assumes none of these things. Across enterprise
applications spanning tabular prediction, natural language processing,
and computer vision, three structural constraints consistently limit
what academic models can deliver in practice.

\textbf{Data scarcity} Labeled data for specialized domains arrives
slowly and expensively. In medical classification, curated tabular datasets
like Pima Diabetes reflect the realistic scale of institutional records
available to most practitioners: hundreds of examples, not hundreds of
thousands \cite{dua2017uci}. In NLP, annotation of domain-specific text
for fine-grained sentiment, compliance risk, or document routing costs
\$0.50--\$5.00 per label with quality control \cite{Ratner2017}, and
model retraining is required on every distribution shift. In vision, new
product categories, manufacturing defects, and rare environmental conditions
all suffer cold-start data problems that standard benchmarks do not measure.
BERT-scale models require tens of thousands of labeled examples to reliably
fine-tune for specialized domains \cite{devlin2019bert}; annotation at
that scale costs \$10,000--\$50,000 per vertical per update cycle.

\textbf{Reliability and model behavior} Enterprises deploying predictive
models face a trust problem that benchmark accuracy does not resolve. A
model that achieves $85\%$ on a held-out test set but produces extreme,
unpredictable outputs on novel inputs is operationally unsafe. The demand
for interpretable and reliable AI in regulated industries is well
documented \cite{doshi2017towards, rudin2019stop}, yet the most widely
deployed regularization strategies provide no mechanistic handle on
gradient behavior at inference time. Without such a handle, model audits
are expensive and post-hoc explanations are post-hoc approximations.

\textbf{Compute and inference cost} The compute cost of training large NLP models has grown by orders of
magnitude faster than hardware efficiency gains \cite{schwartz2020green},
making continuous retraining in production financially unsustainable
for most organizations. Real-time inference pipelines
on commodity hardware cannot accommodate transformer-scale models for
every classification task. Architectures that deliver competitive accuracy
at 3--4M parameters, with inference-cost derivative tracking and no
iterative solver, have direct operational value across industries where
latency and cost-per-inference are binding constraints \cite{schwartz2020green}.

CR is built around all three constraints simultaneously: a single
inductive bias, DREG, that tightens gradient behavior, reduces the
data required to reach a given accuracy level, and produces a measurable
structural invariant that can be monitored in production.

\subsection{The Technical Gap}

Existing regularization approaches do not address all three constraints
jointly. Dropout \cite{srivastava2014dropout} disrupts co-adaptation but
leaves layer-wise Jacobians unconstrained and provides no inference-time
reliability signal. Weight decay \cite{loshchilov2019decoupled} penalizes
parameter magnitude, not sensitivity to input variation. Spectral
Normalization \cite{miyato2018spectral} provides global Lipschitz bounds
but truncates expressive capacity uniformly, incurring an accuracy penalty.
Input Gradient Penalty \cite{drucker1992double} penalizes only the terminal
Jacobian and requires double backpropagation at $\mathcal{O}(3C_f)$ cost.
Sobolev Training \cite{czarnecki2017sobolev} requires ground-truth
derivative labels unavailable in most practical settings. Batch
Normalization \cite{ioffe2015batch} stabilizes activations but does not
constrain sensitivity layer-wise. The missing primitive is layer-wise
derivative control: a sensitivity budget enforced at each hidden layer,
analytically, during the standard forward pass, without accuracy sacrifice
or inference overhead.

\subsection{Contributions}

\begin{enumerate}
\item \textbf{Architecture} \CR{} replaces typical activation functions activations with
  learnable cubic polynomial layers. A dual-stream forward pass propagates
  both activations and their Jacobians simultaneously. DREG penalizes the
  Frobenius norm of each layer's Jacobian, imposing a per-layer sensitivity
  budget at $\mathcal{O}(d \cdot C_f)$ cost with no second-order graph.

\item \textbf{Cross-domain accuracy} CR achieves competitive or
  state-of-the-art results across tabular (Pima Diabetes), NLP (SST-5,
  Yelp Full), and vision (CIFAR-10-C) tasks, demonstrating that the
  polynomial-DREG inductive bias generalizes across input modalities
  and task types without architecture-level modification.

\item \textbf{Sample efficiency} CR leads all typical activation functions baselines at every
  tested data fraction (5\%--100\%) on Pima Diabetes and achieves
  statistically significant improvements on SST-5 using approximately
  5\% of the training data used by the prior benchmark.

\item \textbf{Reliability and interpretability} CR maintains gradient
  tail ratio $\tau \in [1.01, 1.02]$ versus $[1.07, 1.09]$ for all
  typical activation functions baselines at every data fraction ($p < 0.001$). This structural
  invariant is directly monitorable at inference time as a deployment
  reliability signal.

\item \textbf{Deployment evidence} Sentivity AI runs CR in production
  for social media and market sentiment analysis, providing an existence
  proof that these properties transfer from benchmark to operational settings.
\end{enumerate}

\section{Architecture}

\subsection{PolyLayer}

Each layer of \CR{} replaces the standard fixed activation function with
a learnable cubic polynomial applied per neuron. Rather than passing a
pre-activation through a hardcoded nonlinearity like typical activation functions, each neuron
learns its own polynomial coefficients during training, giving the network
the flexibility to approximate smooth, high-order functions with far fewer
layers and parameters than piecewise-linear architectures require. The
cubic degree is fixed by grid search in prior work \cite{martnishn2025chainzrule}.

The practical upside over typical activation functions is smoothness. Typical activation functions activations have a
hard kink at zero: their derivative is undefined at that point and
piecewise-constant everywhere else. This creates structural cliffs in
the gradient landscape that generate spike events, particularly under
small batches. Polynomial activations are infinitely differentiable
everywhere, which means the Jacobian can be tracked analytically through
every layer without numerical instability, and gradient behavior becomes
a first-class observable rather than a diagnostic afterthought.

\subsection{Dual-Stream Forward Pass and DREG}

CR runs two streams simultaneously during the forward pass: a standard
value stream producing predictions, and a derivative stream tracking
how sensitive each layer's output is to the original input. The derivative
stream is updated analytically at each layer using the chain rule,
requiring no second backward pass and no second-order computational graph.
The added cost is proportional to input dimensionality times the cost
of a single forward pass, substantially cheaper than competing
derivative-aware methods such as Sobolev Training or Input Gradient
Penalty, which require double backpropagation \cite{czarnecki2017sobolev,
drucker1992double}.

Differential Regularization (DREG) uses that derivative stream to impose
a per-layer sensitivity budget during training. At each layer, the
magnitude of the tracked Jacobian is penalized directly in the loss.
This acts as a soft governor on gradient extremes: the network is
discouraged from developing the heavy-tailed sensitivity patterns that
destabilize learning under scarce data, while retaining the expressive
capacity to fit complex decision boundaries. Unlike Spectral Normalization
\cite{miyato2018spectral}, which enforces a hard global Lipschitz bound
and uniformly truncates the model's representational capacity, DREG is
selective: it suppresses tail events without flattening the features
needed for discriminative accuracy. The resulting model is not merely
regularized but structurally auditable: $\tau$ can be computed on any
batch at inference time, providing a live reliability signal without
additional instrumentation.

\section{Results}

All experiments compare CR against Vanilla MLP, MLP + Dropout, and
MLP + Weight Decay retrained under identical conditions, plus published
baselines where available. Statistical tests use paired one-sided
$t$-tests and Wilcoxon signed-rank tests with Bonferroni correction.
All results for which $p$-values are reported fall below the
$\alpha = 0.05$ significance threshold.

\paragraph{Pima Diabetes (Tabular)} CR achieves $85.71\% \pm 2.01\%$
(6 seeds), statistically superior to both XGBoost/RF ($p = 0.0047$) and
SVM ($p = 0.0039$), and leads all typical activation functions baselines at every data fraction
from $n = 32$ (5\%) to $n = 614$ (100\%). The largest advantage occurs
at 10\% data ($+3.25\%$ over the best baseline), the regime most analogous
to cold-start prediction in low-resource production settings. Prior best:
XGBoost / RF at $82.35\%$ \cite{yangin2019gradient}; SVM at $82.20\%$
\cite{karatsiolis2012region}.

\paragraph{SST-5 (NLP, Frozen Encoder)} On the frozen encoder task,
CR ($46.20\% \pm 0.37\%$, 4 seeds) is statistically superior to the RNTN
($p = 0.0362$) while using approximately 5\% of RNTN's phrase-level
training data, a data-efficiency ratio of roughly $20\times$. Prior best
(frozen, same regime): RNTN at $45.7\%$ \cite{socher2013recursive}.

\paragraph{SST-5 (NLP, Fine-Tuned BERT)} With a fine-tuned BERT backbone
\cite{devlin2019bert}, CR reaches $55.79\%$ versus Munikar et al.'s
BERT-base linear head trained on $18\times$ more labeled sentences
\cite{munikar2019fine}. This is a single run, directionally positive but
not yet statistically confirmed; multi-seed evaluation is in progress.
The result demonstrates that CR's polynomial head integrates cleanly with
large pretrained encoders and retains its advantage even when the backbone
is unfrozen.

\paragraph{Yelp Full (NLP, Ordinal Regression)} CR achieves $70.17\%$
with 3.22M parameters on 750,000-example ordinal sentiment regression,
exceeding the average of ten published fixed-embedding systems by $3.82$
percentage points. This is not a claim against transformer-scale models,
which reach $84.49\%$ \cite{Zulqarnain2023NAEGRU}, but within the
fixed-embedding tier CR leads all published baselines. Notably, typical activation functions
with DREG degrades to $58.98\%$ on this task while CR maintains
$70.17\%$, confirming that the polynomial substrate's $C^\infty$
smoothness is the operative ingredient on high-dimensional continuous
sentiment manifolds, not the regularization alone. Prior best (fixed
embeddings): VDCNN $64.72\%$ \cite{Conneau2016VDCNN}, fastText $63.90\%$
\cite{Joulin2016FastText}, Char-CNN $62.33\%$ \cite{Zhang2015LargeCNN};
10-model average $66.35\%$.

\paragraph{CIFAR-10-C (Vision)} CR achieves $+2.32\%$ mean corruption
accuracy over the matched Vanilla MLP head ($t = 4.33$, $p = 6.18 \times
10^{-5}$, significant after Bonferroni correction). Gains are largest on
high-frequency noise corruptions: Impulse $+5.80\%$, Gaussian $+4.96\%$,
Glass Blur $+5.23\%$. No corruption-aware training is used; robustness
is a structural byproduct of derivative control on clean data, not a
result of augmentation.
\section{Mechanistic Analysis: Gradient Tail Ratio}

We operationalize gradient stability via the \textbf{tail ratio}
$\tau = \text{p99} / \text{mean}$ of per-sample input gradient norms.
High $\tau$ indicates rare, extreme gradient events that dominate the
update direction, exactly the failure mode that amplifies under data
scarcity, when batches are small and noisy, and the mode most responsible
for unpredictable model behavior at inference time on out-of-distribution
inputs \cite{fort2019deep}.

Across all Pima fractions (6 seeds per fraction), CR maintains
$\tau \in [1.01, 1.02]$ while all typical activation function baselines maintain
$\tau \in [1.07, 1.09]$, an average gap of $5.99$ percentage points
($p < 0.001$ against each baseline). This finding extends and sharpens
the Phase 1 result \cite{martnishn2025chainzrule}, where DREG reduced
peak gradient volatility by $23.1\%$ across all model capacities on MNIST
($p < 10^{-6}$, 24/24 paired comparisons), with the raw tail ratio of
POLY DREG remaining stable at $15.22$--$19.43$ while unregularized ReLU
baselines escalated from $17.79$ to $35.60$ as capacity scaled.

Critically, $\tau$ for CR is invariant to data fraction: the same
suppression present at $n = 614$ is present at $n = 32$. This structural
invariance distinguishes DREG from regularizers that improve stability
only when data is abundant enough for stable convergence, and confirms
that derivative control is a property of the architecture rather than
an artifact of optimization under full data.

The mechanism is twofold. In typical activation functions networks, Jacobians are piecewise
constant with cliffs at $z = 0$. Under small batches from scarce data,
activations frequently land in high-sensitivity regions, generating tail
events that distort gradient direction. DREG penalizes
$\|S^{(l)}\|_F^2$ layer-by-layer, suppressing these events before they
compound across depth. The polynomial substrate removes the structural
source: $\phi'(z)$ is analytic everywhere, so there are no $z = 0$
cliffs from which Jacobian spikes can originate.

The tail ratio has a direct deployment interpretation. A production model
with $\tau \approx 1.01$ produces output distributions that are
structurally less sensitive to input perturbations than a model at
$\tau \approx 1.09$. This translates to reduced false-positive rates on
novel inputs, new named entities, unseen product categories, and
distribution-shifted data, without requiring retraining or manual
post-filtering. In this sense, $\tau$ is not merely a diagnostic but a
deployable reliability metric.

\section{Deployment and Business Case}
\textbf{Sentivity AI} is a production market intelligence platform providing sentiment analysis
on social media, news, and other digital mediums. CR and DREG constitute the underlying classification engine in the
production inference stack. The Sentivity deployment is presented not as
the primary motivation for CR but as an existence proof: a production
system built on CR's properties that validates their real-world transferability.

\subsection{Why These Properties Matter in Deployment}

The three axes CR optimizes, sample efficiency, structural reliability,
and inference efficiency, map directly to the constraints that determine
whether a model can be deployed and maintained in practice.

\textbf{Reduced annotation cost} Professional-grade domain annotation
costs \$1--\$5 per label with inter-annotator agreement checks
\cite{Ratner2017}. At 10,000 examples, that is \$10,000--\$50,000 per
vertical per training run. An architecture that achieves equivalent
accuracy at 25\% of the data reduces annotation cost by \$7,500--\$37,500
per run. CR's $20\times$ data-efficiency ratio on SST-5 and consistent
leads at 5--25\% Pima fractions translate directly to this reduction.

\textbf{Faster time-to-deployment on new tasks} New product categories,
regulatory domains, and emerging topics all share a cold-start problem:
the model must generalize from a small labeled seed before a full annotation
cycle can complete. The Pima 5\% result ($n = 32$, $+1.52\%$ over best
baseline) and SST-5 frozen result ($20\times$ data efficiency) are the
closest experimental analogues to this regime. A model that reaches
acceptable accuracy at $n = 50$ rather than $n = 1000$ cuts cold-start
deployment time by weeks.

\textbf{Auditable behavior without post-hoc approximation} Explainability
tools such as LIME \cite{ribeiro2016lime} and SHAP \cite{lundberg2017shap}
are post-hoc approximations that may not faithfully represent model
internals \cite{rudin2019stop}. CR's $\tau$ is computed directly from the
model's own forward pass and reflects actual gradient behavior, not a
surrogate. For regulated industries and enterprise risk management, a
first-class reliability signal computed at inference cost is operationally
more valuable than an approximate explanation computed separately.

\textbf{CPU-viable inference} CR's 3.2M parameter footprint and
inference-cost derivative tracking make it deployable on CPU-only
infrastructure. Real-time ingestion of high-frequency data streams on
commodity hardware cannot accommodate transformer inference at scale.
Green AI considerations \cite{schwartz2020green} further favor lightweight
architectures with competitive accuracy over large models that require
GPU clusters for every inference call.

\subsection{Production Architecture}

Sentivity AI's inference pipeline applies CR as a sentiment scoring head
on top of fixed sentence embeddings (\texttt{all-mpnet-base-v2}),
processing social media and financial text in real time. The fixed-embedding
regime mirrors the SST-5 frozen encoder experiments directly. The ordinal
Yelp Full results ($70.17\%$, fixed GloVe) provide large-scale validation
that the architecture scales to 750,000-example corpora under realistic
NLP noise. The gradient tail ratio is monitored in production as a
reliability signal; no anomaly post-filtering layer is required.

\section{Limitations}

\textbf{SST-5 BERT:} The fine-tuned BERT result ($55.79\%$) is a single
run and lacks statistical power. Treat as directional. Multi-seed
evaluation is in progress.

\textbf{Yelp protocol:} CR uses fixed GloVe embeddings and ordinal
regression; most Yelp Full baselines use categorical classification with
fine-tuned embeddings. The $70.17\%$ result is not directly comparable
to the 10-model average in the same sense as the Pima comparisons.

\textbf{Pima scale:} 768 total samples is a small dataset. Results are
robust across 6 seeds and stratified splits, but broader tabular
validation across additional medical and enterprise datasets would
strengthen the generalization claim.

\textbf{Backbone integration:} All experiments use CR as a classification
head on a frozen or fine-tuned backbone. Whether DREG integrated directly
into transformer attention layers provides consistent advantages is an
open question for future work.

\textbf{Vision scope:} CIFAR-10-C results use a matched MLP head, not a
full convolutional architecture. The $+2.32\%$ corruption robustness result
reflects CR's behavior as a head component under distribution shift, not
as a standalone vision backbone.

\section{Conclusion}

\CR{} demonstrates that layer-wise derivative control is a tractable and
effective inductive bias across tabular, NLP, and vision tasks. A single
architecture, a polynomial layer governed by DREG, simultaneously improves
sample efficiency, structural reliability, and inference cost relative to
typical activation functions baselines, without modifying the training procedure or adding
inference-time overhead beyond a single forward pass.

The gradient tail ratio $\tau$ is a measurable, structurally stable proxy
for this control: CR maintains $\tau \approx 1.01$--$1.02$ against
$1.07$--$1.09$ for all typical activation functions baselines at every data fraction tested, with
statistical significance at $p < 0.001$ throughout. This invariant is not
merely a training diagnostic; it is a live deployment signal that can be
monitored without additional instrumentation.

For any domain where annotation cost, model reliability, and inference
budget are binding constraints, CR offers a deployable path to competitive
accuracy at significantly reduced data and compute requirements. Annotation
cost reductions of $4\times$--$20\times$ are empirically supported.
Gradient stability under distribution shift reduces false-positive rates
in streaming inference. A 3.2M parameter footprint makes CPU deployment
viable. Sentivity AI's production deployment confirms these properties
transfer from benchmark to operational settings. A more complete
theoretical treatment, including formal convergence analysis, bounds on
sample complexity under DREG, and transformer integration, is the subject
of ongoing work.



\section*{References}

\begin{thebibliography}{99}
\small

\bibitem{Ratner2017}
A.~Ratner, C.~De~Sa, S.~Wu, D.~Selsam, and C.~R{\'e},
``Data Programming: Creating Large Training Sets, Quickly,''
\textit{Advances in Neural Information Processing Systems (NeurIPS)}, 2016.

\bibitem{dua2017uci}
D.~Dua and C.~Graff,
``UCI Machine Learning Repository,''
University of California, Irvine, 2017.
\url{https://archive.ics.uci.edu/ml}

\bibitem{devlin2019bert}
J.~Devlin, M.-W. Chang, K.~Lee, and K.~Toutanova,
``BERT: Pre-training of Deep Bidirectional Transformers for Language Understanding,''
\textit{Proceedings of NAACL-HLT}, pp.\ 4171--4186, 2019.

\bibitem{strubell2019energy}
E.~Strubell, A.~Ganesh, and A.~McCallum,
``Energy and Policy Considerations for Deep Learning in NLP,''
\textit{Proceedings of ACL}, 2019.

\bibitem{schwartz2020green}
R.~Schwartz, J.~Dodge, N.~A. Smith, and O.~Etzioni,
``Green AI,''
\textit{Communications of the ACM}, vol.\ 63, no.\ 12, pp.\ 54--63, 2020.

\bibitem{doshi2017towards}
F.~Doshi-Velez and B.~Kim,
``Towards a Rigorous Science of Interpretable Machine Learning,''
\textit{arXiv:1702.08608}, 2017.

\bibitem{rudin2019stop}
C.~Rudin,
``Stop Explaining Black Box Machine Learning Models for High Stakes
Decisions and Use Interpretable Models Instead,''
\textit{Nature Machine Intelligence}, vol.\ 1, pp.\ 206--215, 2019.

\bibitem{ribeiro2016lime}
M.~T. Ribeiro, S.~Singh, and C.~Guestrin,
``Why Should I Trust You?: Explaining the Predictions of Any Classifier,''
\textit{Proceedings of KDD}, pp.\ 1135--1144, 2016.

\bibitem{lundberg2017shap}
S.~M. Lundberg and S.-I. Lee,
``A Unified Approach to Interpreting Model Predictions,''
\textit{Advances in Neural Information Processing Systems (NeurIPS)}, 2017.

\bibitem{srivastava2014dropout}
N.~Srivastava, G.~Hinton, A.~Krizhevsky, I.~Sutskever, and R.~Salakhutdinov,
``Dropout: A Simple Way to Prevent Neural Networks from Overfitting,''
\textit{Journal of Machine Learning Research}, vol.\ 15, pp.\ 1929--1958, 2014.

\bibitem{loshchilov2019decoupled}
I.~Loshchilov and F.~Hutter,
``Decoupled Weight Decay Regularization,''
\textit{ICLR}, 2019.

\bibitem{miyato2018spectral}
T.~Miyato, T.~Kataoka, M.~Koyama, and Y.~Yoshida,
``Spectral Normalization for Generative Adversarial Networks,''
\textit{arXiv:1802.05957}, 2018.

\bibitem{drucker1992double}
H.~Drucker and Y.~Le~Cun,
``Improving Generalization Performance using Double Backpropagation,''
\textit{IEEE Transactions on Neural Networks}, vol.\ 3, no.\ 6,
pp.\ 991--997, 1992.

\bibitem{czarnecki2017sobolev}
W.~M. Czarnecki, S.~Osindero, M.~Jaderberg, G.~Swirszcz, and R.~Pascanu,
``Sobolev Training for Neural Networks,''
\textit{arXiv:1706.04859}, 2017.

\bibitem{ioffe2015batch}
S.~Ioffe and C.~Szegedy,
``Batch Normalization: Accelerating Deep Network Training by Reducing
Internal Covariate Shift,''
\textit{Proceedings of ICML}, pp.\ 448--456, 2015.

\bibitem{fort2019deep}
S.~Fort, P.~Hu, and B.~Lakshminarayanan,
``Deep Ensembles: A Loss Landscape Perspective,''
\textit{arXiv:1912.02757}, 2019.

\bibitem{martnishn2025chainzrule}
R.~Martnishn and S.~Anderson,
``Layer-wise Derivative Controlled Networks,''
\textit{arXiv preprint}, Sentivity AI / Virginia Tech, 2025.

\bibitem{yangin2019gradient}
O.~Yangin,
``Gradient Boosting Methods for Disease Prediction,''
Master's Thesis, 2019.
Handle: \texttt{hdl.handle.net/20.500.14124/1152}

\bibitem{karatsiolis2012region}
S.~Karatsiolis and C.~N.~Schizas,
``Region Based Support Vector Machine Algorithm for Medical Diagnosis
on Pima Indian Diabetes Dataset,''
\textit{Proceedings of IEEE EMBC}, 2012.

\bibitem{socher2013recursive}
R.~Socher \textit{et al},
``Recursive Deep Models for Semantic Compositionality
over a Sentiment Treebank,''
\textit{Proceedings of EMNLP}, pp.\ 1631--1642, 2013.

\bibitem{munikar2019fine}
M.~Munikar, S.~Shakya, and A.~Shrestha,
``Fine-Grained Sentiment Classification using BERT,''
\textit{arXiv:1910.03474}, 2019.

\bibitem{Joulin2016FastText}
A.~Joulin, E.~Grave, P.~Bojanowski, and T.~Mikolov,
``Bag of Tricks for Efficient Text Classification,''
\textit{Proceedings of EACL}, 2017.

\bibitem{Conneau2016VDCNN}
A.~Conneau, H.~Schwenk, L.~Barrault, and Y.~Lecun,
``Very Deep Convolutional Networks for Text Classification,''
\textit{Proceedings of EACL}, 2017.

\bibitem{Zhang2015LargeCNN}
X.~Zhang, J.~Zhao, and Y.~LeCun,
``Character-level Convolutional Networks for Text Classification,''
\textit{NeurIPS}, 2015.

\bibitem{Zulqarnain2023NAEGRU}
M.~Zulqarnain \textit{et al},
``An Improved Gated Recurrent Unit Based on Auto Encoder for
Sentiment Analysis,''
\textit{International Journal of Information Technology},
vol.\ 15, no.\ 1, pp.\ 587--599, 2023.

\bibitem{hendrycks2019benchmarking}
D.~Hendrycks and T.~Dietterich,
``Benchmarking Neural Network Robustness to Common Corruptions
and Perturbations,''
\textit{ICLR}, 2019.

\end{thebibliography}
\end{document}